\pdfoutput=1
\documentclass[11pt]{article}

\usepackage[]{acl}

\usepackage{times}
\usepackage{latexsym}
\usepackage{multirow}
\usepackage[T1]{fontenc}
\usepackage[utf8]{inputenc}

\usepackage{microtype}

\title{MorisienMT: A Dataset for Mauritian Creole Machine Translation}

\author{Raj Dabre \\
  NICT, Japan \\
  \texttt{raj.dabre@nict.gp.jp} \\\And
  Aneerav Sukhoo \\
  University of Mauritius, Mauritius \\
  \texttt{aneeravsukhoo@yahoo.com} \\}

\begin{document}
\maketitle
\begin{abstract}
In this paper, we describe MorisienMT, a dataset for benchmarking machine translation quality of Mauritian Creole. Mauritian Creole (Morisien) is the lingua franca of the Republic of Mauritius and is a French-based creole language. MorisienMT consists of a parallel corpus between English and Morisien, French and Morisien and a monolingual corpus for Morisien. We first give an overview of Morisien and then describe the steps taken to create the corpora and, from it, the training and evaluation splits. Thereafter, we establish a variety of baseline models using the created parallel corpora as well as large French--English corpora for transfer learning. We release\footnote{\url{https://huggingface.co/datasets/prajdabre/MorisienMT}} our datasets publicly for research purposes and hope that this spurs research for Morisien machine translation. 
\end{abstract}

\section{Introduction}
Neural machine translation (NMT) \cite{bahdanau15} is an end-to-end approach which is known to give state-of-the-art results for a variety of language pairs. NMT, being resource hungry, gives high quality performance for widely spoken resource-rich languages such as English, French, German etc. On the other hand, most languages are resource-poor such as, but not limited to, the vast majority of Indian, African and South-East Asian languages, have to rely on transfer learning either via multilingualism \cite{dabre2020} or monolingual corpora \cite{sennrich-haddow-birch:2016:P16-11} for decent translation quality. Without publicly available datasets, however, it is impossible to develop, let alone evaluate, machine translation for any language. This paper focuses on one such language, Mauritian Creole or Morisien, which is widely spoken in the republic of Mauritius by approximately 1.2 million people.

Creoles are natural languages that develop from the simplifying and mixing of different languages into a new one within a fairly brief period of time. Pidgins, which are simple means of communication between people speaking different languages, typically evolve into creoles. Most creoles are highly related to a widely spoken language, and we focus on Morisien which is a French based creole. Morisien is an important language from the perspective of tourism because Mauritius is a country well known for its tourism industry. Therefore, enabling tourists and locals to easily interact with each other without having to focus on learning each other's languages might help enhance the tourism industry, in addition to enabling better communication between peoples belonging to different nationalities and cultures. For now, we consider it sufficient to focus on translation between Morisien, English and French. 

Although research has been conducted on Morisien in the past \cite{dabre-etal-2014-anou}, there are no publicly available datasets for evaluating machine translation for Morisien--English and Morisien--French. Furthermore, the evaluation was not conducted in a principled manner and the experiments are rather outdated, focusing only on SMT, given that NMT did not exist at the time. Work by \citet{Boodeea2020} is more relevant given current research trends, but they too do not release their datasets, making it difficult to reproduce their work. To this end, we focus more on creating and releasing a dataset with standardized evaluation sets for both language pairs. We first give an overview of Morisien followed by the description of the dataset creation process. We then establish strong baselines using the created parallel corpora, as well as with the use of large helping corpora for French and English. By leveraging transfer learning, we can obtain a translation quality of about 22-22 BLEU for Morisien--English and about 18-19 BLEU for Morisien--French. We also analyze a few examples to show that the translations are indeed of high quality. Our results show that there is significant room for innovation for Morisien NMT and Morisien NLP in general.

\section{Related Work}
This paper mainly focuses on the creation of datasets for under resourced languages, specifically creoles, as well as leveraging transfer learning to improve translation quality.

Recently, there has been significant focus on the curation of data for extremely low-resource languages which are not as widely spoken as some others like English, French, Hindi, etc. In particular, the Masakhane\footnote{\url{https://www.masakhane.io/}} community heavily focuses on African language NLP \cite{nekoto-etal-2020-participatory}, which are numerous but only a few among them are considered as resource rich. Mauritius is considered as a part of Africa, East Africa to be specific, and MorisienMT falls under the broad area of research focusing on African language machine translation. 

Morisien, being a creole, implies that MorisienMT is strongly related to work on creoles \cite{lent-etal-2021-language}. With regard to machine translation, Haitian creole was the first creole language to receive substantial attention \cite{lewis-2010-haitian} and was featured in a WMT shared task\footnote{\url{https://www.statmt.org/wmt11/featured-translation-task.html}}. Work on Morisien itself was focused on a bit later by \citet{Sukhoo2014TranslationBE} and \citet{dabre-etal-2014-anou} but they did not release their datasets. Morisien machine translation was also explored more recently by \citet{Boodeea2020} who trained NMT models, but their datasets were not made publicly available. Motivated by work on Cree \cite{teodorescu-etal-2022-cree}, we decided to focus on the creation of publicly available standardized datasets for Morisien to/from English and French translation.

Morisien is a low-resource language, and that low-resource settings are often supplemented with transfer learning. In particular, transfer learning approaches such as pre-training followed by fine-tuning \cite{zoph-etal-2016-transfer} are most relevant. Multilingual training approaches \cite{dabre2020,firat16} may also be leveraged, but given the skew in the corpora sizes for resource-rich pairs and pairs involving Morisien, the fine-tuning paradigm is more relevant. More recent approaches involving self-supervised pre-training such as mBART are also attractive but given that the monolingual corpus for Morisien is rather tiny, if not non-existent, focusing on crawling monolingual corpora for Morisien will need to be prioritized before self-supervised pre-training can be leveraged.

\section{Morisien}
Mauritian Creole, also known as Morisien, is spoken in Mauritius and Rodrigues islands.  A variant of Morisien is also spoken in Seychelles. Mauritius was colonized successively by the Dutch, French and British. Although the British took over the island from the French in the early 1800, French remained as a dominant language and as such Morisien shares many features with French.

\begin{table}[]
\centering
\begin{tabular}{c|c|c}
\textbf{French} & \textbf{Morisien} & \textbf{English} \\\hline
avion           & avion           & airplane        \\
bon             & bon             & good             \\
gaz             & gaz             & gas              \\
bref            & bref            & brief            \\
pion            & pion            & pawn            
\end{tabular}
\caption{Similarities between French and Morisien.}
\label{tab:frcr}
\end{table}

\begin{table}[]
\centering
\begin{tabular}{c|c|c}
\textbf{French} & \textbf{Morisien} & \textbf{English}          \\\hline
mauvais         & move              & move                      \\
confort         & konfor            & comfort                   \\
méditation      & meditasion        & meditation                \\
insecte         & insekt            & insect                    \\
condition       & kondision         & state, terms \\
                &                   &
or provision
\end{tabular}
\caption{Differences in accent usage between Morisien and French.}
\label{tab:frcraccents}
\end{table}

\subsection{Morisien--French Similarities}
The same alphabets are used in both cases, and they are pronounced similarly.  In addition, some words are written and pronounced in the same way.  Table~\ref{tab:frcr} contains some examples. Furthermore, in written French there is a heavy usage of accents which is absent in Morisien. Many words are pronounced similarly in French and Morisien, but the way they are written is different. Some examples are given in Table~\ref{tab:frcraccents}.

\subsection{Morisien Grammar}
The grammar of Morisien has been published in 2011 by Daniella Police-Michel in the book Gramer Kreol Morisien\footnote{\url{https://education.govmu.org/Documents/educationsector/Documents/GRAMER\%20KREOL\%20MORISIEN\%202211.pdf}}.Morisien sentence structure follows the subject-verb-object order, the same as English and French. However, some similarities and differences with English and French can be noted as follows:

\begin{enumerate}
    \item Like French but unlike English, adjectives are sometimes placed after the object rather than before. ``The brown bird'' is translated as: ``Zwazo maron-la''. Here, ``maron'' stands for ``brown'' and is moved after the object (Zwazo). The article “la” which stands for “the” is moved at the end of the sentence. On the other hand, the French translation would be ``L’oiseau maron'' which shows that Morisien is more grammatically similar to French in terms of adjective placement but differs in terms of article placement.
    \item Singular and plural forms are different between English and Morisien. ``There are many birds'' is translated as ``Ena boukou zwazo'' where the plural form ``zwazo'' does not take the suffix ``s'' as in English. Instead, the word ``boukou'' indicates ``many'' and therefore, it can be deduced that there are many birds.	In French, the translated sentence is ``Il y a beaucoup d’oiseaux'' which has the same grammatical construction as its Morisien equivalent.
    \item Verbs are sometimes dropped in Morisien. ``He is bad'' is translated as ``Li move'' where ``He'' is translated to ``Li'' and ``bad'' to ``move''.  The verb ``is'' is dropped. Furthermore, in French, the translated sentence becomes “Il est méchant”, where the verb is retained, indicating a difference from Morisien.
\end{enumerate}

\section{MorisienMT}
The data for MorisienMT was created manually, specifically through books available in English translated to Morisien and French. One such source was the holy Bible. We also created basic sentences and useful expressions manually from scratch for all 3 languages. Not all English content is translated into both languages, and thus there is more Morisien--English data than Morisien--French data. There is also a small amount of monolingual corpus which we extracted from various sources. Most of the aforementioned data is similar to the one used by \citet{dabre-etal-2014-anou} but we noticed that there were several issues in the version of the data they used such as non-standard splits and improper punctuation.

\subsection{Dataset Cleaning}
Upon manual investigation of the dataset from, \citet{dabre-etal-2014-anou} we found that the sentences were of reasonably high quality, owing to being translated by a native speaker. However, a major problem we observed was improper punctuation. We found that spaces were inserted before full-stops, question marks and commas inconsistently. Additionally, in French, words like ``l'homme'' were sometimes written as ``l' homme''. We used regular expression matching and fixed all these issues. We did not discard any content, and we ended up with 23,310 and 16,739 pairs for English--Morisien and French--Morisien.

\subsection{Evaluation Splits}
Of the 23,310 pairs for English--Morisien, 12,467 were dictionary entries. Similarly, for French--Morisien, of 16,739 pairs 12,424 were dictionary entries. Since the main goal is to develop translation systems that can translate full sentences, we decided to choose the longest sentences for the development and test sets. Furthermore, we decided to have trilingual evaluation sets following \citet{guzman2019flores} and \citet{goyal2021flores101}. To this end, we first used Morisien as a pivot and extracted a trilingual corpus of 13,861 sentences. Next, we sorted the corpora according to the number of words on the Morisien side and chose the top 1,500 ones representing the longest sentences. We then randomly chose 500 for the development set and 1,000 for the test set, both of which are trilingual. This is the major difference between previous works and ours, since our evaluation set is intended to focus on 1,500 proper sentences. We remove the pairs from the English--Morisien and French--Morisien corpora that overlap with the development and test set, resulting in 21,810 and 15,239 pairs respectively. We also filter the monolingual corpus and end up with 45,364 sentences. 

Table~\ref{tab:corporastats} contains an overview of the corpora. It is evident that there is a big mismatch between the length distributions of training and evaluation sets, but we prioritize the evaluation of medium to longer length sentences, we have no other choice. However, from the results in Section~\ref{sec:results} it will be evident that even when the training data contains mostly dictionaries, we can obtain a fairly high translation quality.

\begin{table}[]
    \centering
    \begin{tabular}{c|c|c|c}
    \hline
        \multicolumn{4}{c}{\textbf{English--Morisien}}\\\hline
        \textbf{split} & \textbf{L} & \textbf{AL-s} & AL-t\\\hline
        train & 21,810 & 6.5 & 5.8 \\
        dev & 500 & 16.9 & 16.2\\
        test & 1,000 & 17.0 & 16.0 \\\hline
        \multicolumn{4}{c}{\textbf{French--Morisien}}\\\hline
        \textbf{split} & \textbf{L} & \textbf{AL-s} & AL-t \\\hline
        train & 15,239 & 2.6 & 2.0 \\
        dev & 500 & 18.0 & 16.2 \\
        test & 1,000 & 18.0 & 16.0 \\\hline
        \multicolumn{4}{c}{\textbf{Morisien Monolingual}}\\\hline
        \textbf{split} & \textbf{L} & \textbf{AL} & -\\\hline
        - & 45,364 & 15.8 & -\\
    \end{tabular}
    \caption{Corpora statistics for MorisienMT. L indicates number of lines, AL indicates average sentence length and -s, -t indicate source or target language.}
    \label{tab:corporastats}
\end{table}

\section{Experiments}
We describe the experimental settings including datasets used, training details, and models.

\subsection{Datasets}
In addition to MorisienMT, we use 5M randomly sampled sentence pairs from the UN corpus for French--English  \cite{ziemski-etal-2016-united} which we use for pre-training a French--English bidirectional model. We use the validation set from the UN corpus for early stopping.

\subsection{Training details}
We train transformer \cite{NIPS2017_7181} models using the YANMTT toolkit \cite{DBLP:journals/corr/abs-2108-11126} which is based on the HuggingFace transformers library. We first create a joint English, French, Morisien sub-word tokenizer using sentencepiece \cite{kudo-richardson-2018-sentencepiece} consisting of 16,000 subwords, which we use for all our experiments and is shared between the encoder and decoder. The training data of the tokenizer comes from the training sets of MorisienMT. We trained baseline models for English--Morisien and French--Morisien by varying hyperparameters such as number of layers, hidden-sizes, dropouts, label-smoothing and learning rates. We find that using the transformer-base architecture \cite{NIPS2017_7181} as it is but choosing dropouts of 0.2, label-smoothing of 0.2 and learning rate of 0.0001 with the ADAM optimizer gave the best results. For pre-training, we use the transformer-big architecture with default hyperparameter values as in \cite{NIPS2017_7181}. Instead of separate unidirectional models, we pre-train a single bidirectional model which translates French and English to the other language. We train this multilingual model using the language indicator token proposed by \citet{johnson-etal-2017-googles}. We then fine-tune the pre-trained models separately for English--Morisien and French--Morisien using the same hyperparameters as for the baseline models without fine-tuning. All models are trained to convergence on the relevant development sets, where convergence is said to take place if the development set BLEU score does not increase for 20 consecutive evaluations. BLEU scores are calculated using sacreBLEU with default parameters \cite{post-2018-call}. 

For decoding, we choose the model checkpoint with the highest validation set BLEU score and use a default beam size of 4 and length penalty of 0.8.

\begin{table*}[t]
    \centering
    \begin{tabular}{c|c|c}
    \multirow{5}{*}{1} & \textbf{Input}     &  Ena mem ki tom lor bann serviter, maltret zot e touy zot. \\
    & \textbf{Reference}     & Others grabbed the servants, then beat them up and killed them. \\\cline{2-3}
    & \multicolumn{2}{c}{\textbf{Translations}}\\\cline{2-3}
    & \textbf{Baseline} & Some have been agreed on those servants, and they are murdered.\\
    & \textbf{Fine-tuned} & Some people even fall on servants, maltreat them and kill them.\\\hline
    
    \multirow{5}{*}{2} & \textbf{Input}     &  “E natirelman mo prezant mo bon kamarad, Murgat”, Madam Urit finn kontinye.  \\
    & \textbf{Reference}     & Mrs Octopus continued, “And naturally, I present my good friend Mr Squid”. \\\cline{2-3}
    & \multicolumn{2}{c}{\textbf{Translations}}\\\cline{2-3}
    & \textbf{Baseline} & “Hey, I’ve got a good friends, Mr Octopus.”\\
    & \textbf{Fine-tuned} & “Hey obviously I present my good friend, Squid”, Mrs Octopus went on. \\\hline
    \end{tabular}
    \caption{Examples for Morisien to English translation.}
    \label{tab:examples}
\end{table*}

\subsection{Models trained}
We train and evaluate models for Morisien to English, English to Morisien, French to Morisien and Morisien to French. For each direction, we have baseline models without pre-training and fine-tuned models.

\begin{table}[]
    \centering
    \begin{tabular}{c|c|c|c|c}
    \multirow{2}{*}{\textbf{Model}}     & \multicolumn{4}{c}{\textbf{Direction}} \\
    \cline{2-5}
    & \textbf{cr-en} &  \textbf{en-cr} & \textbf{cr-fr} & \textbf{fr-cr}\\\hline
    \textbf{Baseline} & 9.1 & 9.9 & 4.6 & 5.6 \\
    \textbf{Fine-tuned} & \textbf{22.9} & \textbf{22.6} & \textbf{17.9} & \textbf{19.2} \\\hline
    \end{tabular}
    \caption{Baseline and fine-tuned model results for translation involving Morisien (cr), English (en) and French (fr). Clearly, fine-tuning leads to substantial gains in all directions.}
    \label{tab:results}
\end{table}
\section{Results}\label{sec:results}
Table~\ref{sec:results} compares models trained from scratch and via fine-tuning for 4 translation directions: Morisien--English, English--Morisien, Morisien--French and French--Morisien. Owing to the tiny training set, most of which is a dictionary, baseline models without any pre-training show poor performance. This is especially the case for translation involving French and Morisien. However, fine-tuning the bidirectional French--English model trained on the UN corpus leads to large improvements. We use only 5 million out of 11 million sentence pairs from the UN corpus, and we expect further gains if the corpus size is increased.

\subsection{Translation examples}
We show in Table~\ref{tab:examples} some translation examples of baseline and fine-tuned models for Morisien to English translation. In the first example, taken from the holy Bible, the baseline system mistakes the act of ``grabbing the servants'' for ``agreeing with the servants'' and misses the part where the ``servants are beaten up''. On the other hand, the fine-tuned model manages to capture both phenomenon properly. Both systems make the mistake of translating ``others'' as ``some'' but this is understandable because a translation of the word ``ena'' in Morisien in English is ``some''. The fine-tuned system also uses the word ``maltreat'' instead of ``beat'' and while this does reduce the adequacy of the translation, the general meaning is conveyed properly.

In the second example, taken from a story book, the baseline system completely mistranslates the Morisien sentence. On the other hand, the fine-tuned model, except for the placement of the phrase ``Mrs Octopus went on'' to the end of the sentence and the imprecise translation of ``natirelman'' to ``obviously'', manages to translate almost perfectly. In the reference, ``Mrs Octopus continued'' is at the beginning of the sentence, and in the translation, ``Mrs Octopus went on'' is at the end of the sentence. The equivalent of ``Mrs Octopus went on'' in Morisien, ``Madam Urit finn kontinye'', is also at the end of the sentence and this explains the positioning in the translation. Multiple references may help in more realistic evaluation by not penalizing such translations.

\section{Conclusion}
We have presented MorisienMT, a dataset for machine translation between Mauritian Creole (Morisien) to/from English and French. Our datasets contain parallel dictionaries and sentence pairs belonging to a mix of domains and their sizes range from roughly 17,000 to 23,000 pairs. We also provide a monolingual corpus for Morisien containing about 45,000 sentences. We conduct translation experiments using MorisienMT in conjunction with large English--French corpora and show the pre-training on larger corpora can yield improvements of up to 13 BLEU. This large improvement, despite the training data containing mostly dictionary entries but evaluation data containing full sentences, shows that there is a possibility of leveraging dictionaries for creoles and pre-trained models for high quality translation. In the future, we plan to expand MorisienMT with additional data as well as on additional generation tasks for Morisien. We have not focused on the use of the Morisien monolingual corpus we have released, and intend to do so in the future after augmenting it substantially.
\bibliography{anthology,custom}
\bibliographystyle{acl_natbib}

\appendix

\end{document}